\PassOptionsToPackage{margin=2.5cm}{geometry}
\documentclass[12pt]{article}
\usepackage[basic]{wordlike}
\usepackage{lipsum}% just to generate filler text
\usepackage{setspace}
\usepackage[pdftex]{graphicx}
\usepackage{url}
\usepackage{multirow}
\usepackage{verbatim}
\usepackage{color}

\usepackage{soul}

\usepackage{titlesec}
\titleformat{\subsection}    
       {\normalfont\fontsize{12}{17}\bfseries\itshape}{\thesubsection}{1em}{}

\begin{document}
\markboth{Instrumentation and Measurement Magazine}%
{Measuring Cognitive Status from Speech}



%\maketitle

\noindent \textbf{Measuring Cognitive Status from Speech in a Smart Home Environment}\\
\noindent \textbf{Kathleen C.\@ Fraser and Majid Komeili}

\bigskip 
\bigskip 

%\IEEEpeerreviewmaketitle


%\begin{abstract}
%The abstract goes here.
%\end{abstract}

%\begin{IEEEkeywords}
%\end{IEEEkeywords}

% \fi
%\IEEEpeerreviewmaketitle

%\section*{Introduction}
%\IEEEPARstart{T}{he} 

\noindent The population is aging, and becoming more tech-savvy.  
The United Nations predicts that by 2050, one in six people in the world will be over age 65 (up from one in 11 in 2019), and this increases to one in four in Europe and Northern America.
Meanwhile, the proportion of American adults over 65 who own a smart phone has risen 24 percentage points from 2013-2017, and the majority have Internet in their homes \cite{anderson2017technology}. Smart devices and smart home technology have profound potential to transform how people age, their ability to live independently in later years, and their interactions with their circle of care. 

Cognitive health is a key component to independence and well-being in old age, and smart homes present many opportunities to measure cognitive status in a continuous, unobtrusive manner.

In this article, we focus on \textit{speech} as a measurement instrument for cognitive health. Existing methods of cognitive assessment suffer from a number of limitations that could be addressed through smart home speech sensing technologies.
We begin with a brief tutorial on measuring cognitive status from speech, including some pointers to useful open-source software toolboxes for the interested reader. We then present an overview of the preliminary results from pilot studies on active and passive smart home speech sensing for the measurement of cognitive health, and conclude with some recommendations and challenge statements for the next wave of work in this area, to help overcome both technical and ethical barriers to success. 

\section*{Measuring Cognitive Decline from Speech}

Declines in cognitive ability can occur as a result of a number of underlying physical factors: traumatic brain injury, stroke, and neurodegenerative diseases such as Alzheimer's disease (AD), among others. The set of cognitive symptoms associated with neurodegenerative diseases is known as \textit{dementia}. Since dementia develops slowly over time, there is increasing interest in methods to detect it at the earliest possible stages, when interventions aimed to slow down the decline process are most effective. It is estimated that approximately half of dementia cases go undiagnosed, which can lead to poorer outcomes for both the individual and their caregivers \cite{lyons2015pervasive}.

Part of the challenge lies in the fact that `cognition' is not a physical property to be measured, and all measures of cognition are by necessity indirect. A physician may look at an MRI image and see the patterns of atrophy in the brain, but in many cases such information is only roughly correlated with the actual behaviour and function of the patient, particularly at the early stages. Typically, cognition is measured through standardized `cognitive tests', which target different areas of cognition and are validated against existing population norms. However, as a measurement instrument, these tests suffer from both practical and theoretical limitations: they are administered in a clinic by a trained professional, which is expensive and time-consuming, but also affects the measurement itself, as social anxiety and pressure to do well can potentially lead to poorer performance. Furthermore, the measurements are not generally repeatable, due to a learning effect, making measurement reliability difficult to estimate. At the same time, repeated measurements are highly desirable, as it is known that people with dementia have ``good days and bad days,'' and cognitive ability can vary from day-to-day. 
%\cite{rockwood2014good}.  
Alternative measurement instruments, particularly those that can be applied outside of the clinic, are needed.

Speech and language offer a natural and incredibly rich view into cognitive function. In dementia, various cognitive areas such as memory, executive function, attention, processing speed, and language itself may be affected, and these deficits can all show up in what a person says and how they say it. Therefore, there has been growing interest in automatically detecting the changes in speech and language that indicate underlying cognitive decline.

Preliminary lab-based studies have supported the hypothesis that speech carries information about cognitive status. For example, a machine learning classifier trained on linguistic and acoustic features extracted from short speech samples was able to distinguish between participants with and without AD with 81\% accuracy \cite{fraser2016linguistic}. Other work used automatic conversation analysis to differentiate between participants with  progressive
neurodegenerative dementia and those with functional memory disorders with 90\% accuracy \cite{mirheidari2016diagnosing}. %anything in IEEE?

Perhaps even more promising is the use of automated speech analysis to detect mild cognitive impairment (MCI), a period of subtle decline which can occur prior to dementia. Roark \textit{et al.} distinguished between patients with MCI and healthy older controls with an area under the receiver operating characteristic curve (AUC) of 0.86 based on measures relating to speech rate and pause rate, as well as syntactic measures~\cite{roark2011spoken}; similar results have subsequently been obtained by other research groups \cite{beltrami2018speech,gosztolya2019identifying}.

\begin{figure*}
    \centering
    \includegraphics[width=14cm]{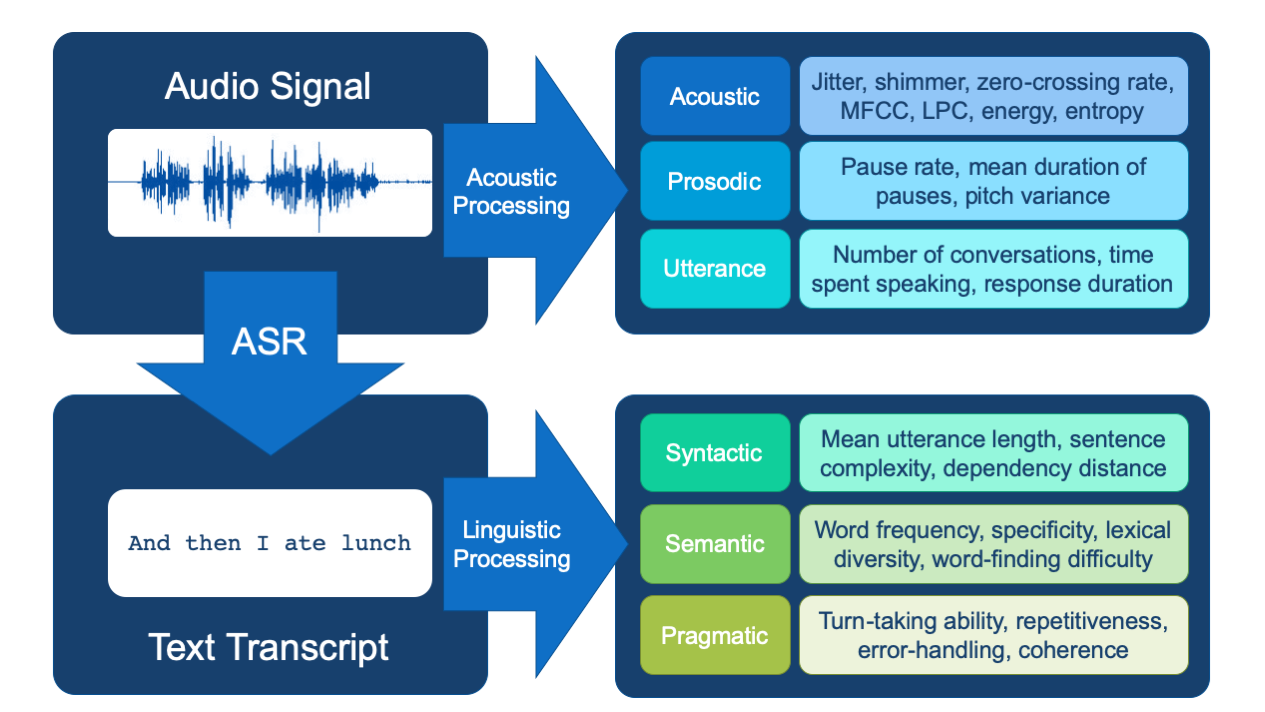}
    \caption{Example features extracted from speech data for the measurement of cognitive status. While a complete description of the features is outside the scope of this article, detailed feature descriptions are available in the literature, e.g., \cite{fraser2016linguistic, mirheidari2016diagnosing, roark2011spoken, beltrami2018speech,komeili2019talk2me, crocco2016audio}. }
    \label{fig:features}
\end{figure*}

We now turn to the practical details involved in speech analysis. Note that the term \textit{speech analysis} actually encompasses a wide range of lower-level features when we delve into the details of the processing. Figure~\ref{fig:features} offers a non-exhaustive summary of the features that can be extracted from speech and linked to cognition. Such features can include low-level acoustic metrics of voice quality and regularity, prosodic features relating to pausing and speech rate, as well as summary features relating to how much time is spent speaking. 
For example, OpenSMILE\footnote{https://www.audeering.com/opensmile/} is an open-source tool that can be used to extract 1582 acoustic features from the Interspeech 2010 Paralinguistic Challenge feature set.

Audio files can then be automatically transcribed using automatic speech recognition (ASR) software, e.g.\@ Kaldi,\footnote{https://kaldi-asr.org/} which is an open-source speech recognition toolkit.
By analyzing the transcript, a wealth of additional information becomes available: features relating to the complexity and grammaticality of utterances  (\textit{syntactic} features), word repetitions, the variety of different words used and their linguistic properties (\textit{semantic} features), and even the speaker's ability to carry on a conversation and handle misunderstandings (\textit{pragmatic} features) are all indicative of cognitive health. The interested reader is referred to \cite{komeili2019talk2me} for a more complete description of the commonly used features listed in Figure~\ref{fig:features}. COVFEFE is an open-source package for extracting approximately two thousand acoustic, prosodic, utterance as well as syntactic, semantic and pragmatic features \cite{komeili2019talk2me}. This package can be applied arbitrarily to studies that include linguistic data.

While individual features could be linked directly to cognitive status, in most current research the features are combined in a machine learning framework. The machine learning models used in the literature vary from simple logistic regression and decision tree classifiers, to state-of-the-art deep learning models. Such models vary in terms of complexity and interpretability, and the choice depends on a number of factors including how much training data is available, the data representation or number of features, and whether the output should be categorical (classification) or continuous (regression). The Python package \texttt{scikit-learn} is one widely-used option for machine learning.\footnote{https://scikit-learn.org/}

Thus, a typical analysis pipeline would involve recording the speech, generating text from the speech, extracting features from the audio and/or text streams, and then feeding the features to a machine learning model to output an estimate of cognitive status. Evidently, there are numerous sources of error and uncertainty along that pipeline that may propagate through the system. In their discussion on uncertainty in medical measurements, Parvis and Vallan enumerate three distinct categories of medical measurement instruments: (1) those that directly measure some physical quantity of clinical interest; (2) those which convert the extracted quantities into more interpretable features; and (3) those which combine several quantities into a single indicator \cite{parvis2002medical}. Clearly, the measurement of cognition from speech falls into the last and most complex category.

Accordingly, there are a number of concerns that arise when we consider the current state-of-the-art in using speech as a measurement instrument for cognition:

\begin{itemize}
    \item \textbf{Reference standard:} In many cases, researchers consider a brief cognitive screen, such as the Mini-Mental Status Exam (MMSE), 
    %\cite{folstein1975mini, tangalos1996mini} 
    to be the reference standard. In addition to the general problems of cognitive testing instruments mentioned above, the MMSE has specific known issues, including non-linearity as well as floor and ceiling effects -- a highly-educated individual, for example, may report a decline in their own cognition while still scoring within the `normal' range on the MMSE. Therefore, engineering a speech-based instrument which faithfully reproduces the MMSE would represent progress, but could not claim to measure the `true' value of cognition, which we must take to be ``not only unknown but also unknowable'' \cite{ferrero2006measurement}.
    
    \item \textbf{Repeatability:} Ideally, if we use the same instrument to measure the same quantity under the same conditions, we will receive the same measurement. However, the vast majority of research on speech-based dementia detection considers only a single speech sample per person, making it impossible to assess the stability of the measurement. When uncertainty is discussed, it is typically only in the context of the machine learning algorithm: if some parameter is changed (e.g., the training data in the case of cross-validation), then what effect does it have on the accuracy of the prediction with respect to the reference? This source of variance in the outcome is sometimes reported in terms of confidence intervals or statistical significance. However, to our knowledge the following basic experiment has not been conducted: If we took a speech sample from a person today and another sample tomorrow, extracted the same features from each and fed them to the same machine learning model, how closely would the outputs match? This concept of repeatability (or in the medical community, \textit{test-retest reliability}) is not assessed in single-sample, cross-sectional study designs. 
    
    \item \textbf{Resolution}: As discussed, many studies characterize cognitive status as a binary classification: participants are labelled as either having dementia, or not. To monitor change over time, it will be necessary to recognize cognition as a continuous measure. Some work has started to move in this direction by re-formulating the classification task as a regression task, with the aim of predicting an MMSE score rather than a diagnostic category. While this would certainly represent an improvement in resolution, even MMSE is a relatively coarse scale from the metrology perspective, with only 30 possible values to cover the entire range of human cognitive ability.
        
    \item \textbf{Sensitivity:} Most studies on speech-based measures of cognition involve participants who have already been diagnosed with dementia, due to the difficulties in identifying and recruiting study participants who \textit{will develop} cognitive impairment over the course of a time-limited study period. However, this limits our understanding of how early in the disease trajectory we can actually pick up on changes. Computational studies of speech from public figures, such as politicians 
    %\cite{berisha2015tracking} 
    and professional athletes \cite{berisha2017longitudinal}, suggest that the changes may be detectable from speech prior to formal diagnosis, but this work has yet to be extended to a large sample of ordinary citizens.

\end{itemize}

Continuous speech sensing in a smart home environment can mitigate some of these issues. While reference to an established clinical instrument such as the MMSE is still necessary, continuous longitudinal sensing makes it possible to use an individual's own normal functioning as a reference, and detect decline with respect to that baseline -- that is, to calibrate the measurement relative to each individual. Furthermore, continuous (or high-frequency) speech recording should make it possible to measure the variance in the signal, develop confidence intervals around the measurement, and detect deviations from that normal range.  Finally, if sensors are installed in the homes of cognitively healthy seniors and allowed to operate unobtrusively over time, then in cases where the user does develop dementia, we will have a better understanding of the sensitivity of speech as a measurement of cognitive decline.  

\section*{Smart Home Speech Sensing}

\begin{figure*}[th]
    \centering
    \includegraphics[width=16cm]{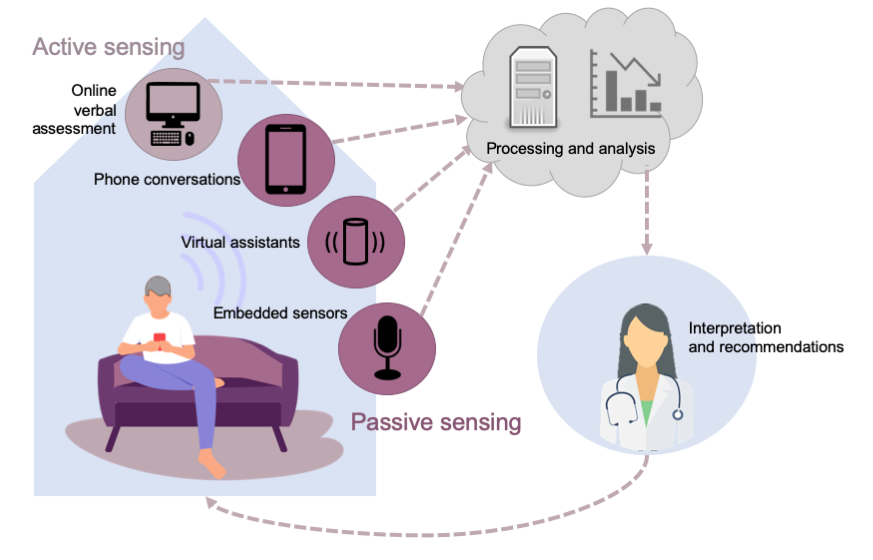}
    \caption{A number of audio sensing options for smart homes have been proposed. Here, arrows show the flow of information (raw data, results of automatic processing, physician-verified health information) through the system.}
    %, including: online assessments recorded over the computer or tablet microphone; an app installed on the user's phone to record telephone conversations; interactions between a user and their smart home virtual assistant (e.g., a device like the Amazon Echo, or even a robot), and embedded microphones to record all ambient audio data from the home. }
    \label{fig:smart_home}
\end{figure*}

There have been a number of proposals for how smart home speech sensing might be implemented (see Figure~\ref{fig:smart_home}). These proposals fall broadly into two categories: \textit{active sensing}, in which users must perform some action to initiate the recording, and \textit{passive sensing}, in which the recording occurs unobtrusively, without the user's intervention.  In either case, the encrypted audio data is then sent to an external server for processing and analysis. If a concerning trend is detected, the user's healthcare provider is notified for further assessment and possible intervention.

\subsection*{Active Sensing}
There has been growing interest in providing mobile cognitive assessments in the form of apps that can be deployed on a tablet or smart phone. Many of these assessments are simply digitized versions of traditional pen-and-paper tests, and are designed to be administered by a health professional, rather than independently in the home. Here we focus instead on a few examples of remote, repeated, home-based assessments where speech was a primary outcome measure. 

Jeffrey Kaye at the Oregon Health and Science University proposed that cognitive tests displayed on a home computer screen, recorded over the microphone, and scored using ASR would greatly improve outcome assessment in clinical trials for AD medications, by allowing assessments to occur frequently, over a longer period of time, in a manner convenient to participants \cite{kaye2008home}. A test system was deployed, along with wireless passive infrared motion sensors, in the homes of 265 older participants for an average of 33 months \cite{kaye2011intelligent}. As part of the protocol, participants took part in daily conversations with research assistants over the computer webcam. Subsequent analyses of the conversations were able to detect MCI based on vocabulary richness measures with AUC = 0.71 \cite{dodge2015social}, and with an accuracy of 84\% based on the psycholinguistic properties of the words used \cite{asgari2017predicting}. In a similar vein, the \textit{CAL: Cognitive Assisted Living} smart home architecture \cite{bowers20183cap}, and the \textit{Lilly Exploratory Digital Assessment Study} of smart device sensors~\cite{chen2019developing} both incorporated regularly-scheduled, audio-recorded interviews for the purposes of detecting cognitive status through speech. 

A potential drawback to these types of active assessments is that they place a burden on the participant to complete, which over time may lead to boredom, apathy, or non-compliance. Depending on the nature of the tasks, they may also suffer from a learning effect, much like traditional cognitive testing. They also assume some level of technological fluency, for the user to log on to the device and complete the procedure \cite{sabbagh2020early}.

\subsection*{Passive Sensing}

Passive sensing frameworks have the benefit of collecting speech data unobtrusively over long periods of time. However, they also generate more technical challenges, as well as ethical issues surrounding privacy and consent, which will be discussed in more detail below. First, we examine some of the proposed passive sensing technologies.

\bigskip 

%\subsubsection{Smartphone conversation recording}

\textbf{Smartphone conversation recording:} As an increasing proportion of the population owns a smart phone, it is a natural choice for recording health data, including variables such as movement, location, heart rate, respiration, and of course voice.  In \cite{netscher2016applications}, conversations were recorded between participants with early dementia and a family caregiver, and acoustic and linguistic processing was used to identify the speakers with dementia with 92\% accuracy. While the conversations were actually recorded live, the author proposed that a similar methodology could be used to analyze phone conversations with family members \cite{netscher2016applications}.  A group from IBM Research recorded regular phone conversations between elderly individuals and a monitoring service who called to check in on their clients at home \cite{shinkawa2018word}. They found that the participants with dementia repeated themselves more from day-to-day than those without dementia. One might suspect that users would object to the idea of their phone calls being recorded and analyzed; however, it appears that even just analyzing the frequency of conversation, without recording the audio itself, is correlated with some measures of mental well-being \cite{kourtis2019digital}.

%\subsubsection{Wearable devices}
\bigskip

\textbf{Wearable devices:} Wearable recording devices, as a category distinct from smart phones, have been considered in a few research studies. %\cite{lyons2015pervasive} describe small digital micro-recorders, which recorded the time and duration of user conversations.
A group at Cornell used a mobile sensing device to capture both physical activity and audio data from older adults in a continuing care retirement community. Acoustic features were extracted from the speech signal without actually recording the raw data, to protect privacy. The proportion of time spent in conversation correlated with mental health scales for social functioning, as well as correlating negatively with a depression scale \cite{rabbi2011passive}.  In another study, researchers used a wearable electronically-activated recorder to intermittently sample audio snippets throughout the day, and found that markers of complex, analytic, and specific language correlated with measures of executive function and working memory \cite{polsinelli2020natural}. 

%% WBANS

%\subsubsection{Smart home virtual assistants}
\bigskip 

\textbf{Smart home virtual assistants:} One exciting possibility for speech sensing in the home is the use of virtual assistants. Virtual assistants (such as Google Home and Amazon Alexa) provide a natural interface to smart home features and assistive technologies, with older adults finding voice interfaces more user-friendly than touch-based tablet interfaces~\cite{kobayashi2019effects}.  Technological advances in natural language processing and machine learning are making virtual assistants more capable. This enables a richer conversation between user and the virtual assistant which in turn could provide a rich source of information for long-term study of cognitive status in a smart home environment.
Since these spoken interactions would be analyzed in any case for the purposes of  question-answering and information retrieval, it would require minimal additional processing to extract features relevant to cognitive status.

With work in this area ongoing, preliminary results have shown promise: researchers found statistically significant correlations between MMSE scores and features measuring pauses, hesitations, and error-handling when older adults interacted with virtual assistants \cite{kobayashi2019effects}; achieved 79\% accuracy in predicting future car accidents (within 1.5 years) based on  pauses, filled pauses, and pronoun use in conversations with virtual assistants \cite{yamada2020predicting}; and  found that people with MCI pause more often and for longer, and produce fewer words and shorter speech chunks when interacting with virtual assistants than do cognitively healthy users \cite{walker2020characterising}.  

In additional to the existing commercial offerings, another type of digital assistant may soon be available: personal care robots. Often proposed as an assistive technology to allow older adults to live in the community longer, voice-controlled robots could also record and analyze the user's voice for signs of cognitive decline. In one study, a conversational robot was found to be generally well-liked and engaging by participants with AD, and a number of the features extracted from the dialogues distinguished between persons with mild, moderate, and severe dementia \cite{pou2020conversational}.

%\subsubsection{Ambient embedded sensors}
\bigskip

\textbf{Ambient embedded sensors:} Perhaps the most comprehensive speech sensing methodology is to embed microphones throughout the smart home and record everything that occurs. Embedded microphone arrays have been previously proposed to help detect falls and to recognize cries for help.  The SWEET-HOME project in France focused intensively on bringing audio technology to smart homes for older people, and built a home lab with 150 sensors and actuators, including multiple omnidirectional microphones. They found that audio recording was much more acceptable to older users than video recording \cite{portet2013design}. The focus of that and other audio-based smart home projects has typically been on providing assistive technologies, not assessing cognitive state, though there is clearly potential for further research. For example, it has been proposed that  ambient assistive living technologies for persons with dementia could additionally report some behavioural metrics to the care team, including instances of repetitive speech \cite{amiribesheli2018tailored}. However, privacy preservation is paramount in the extensive recording scenarios discussed in this section.

\section*{Challenges}

\begin{figure*}
    \centering
    \includegraphics[width=12cm]{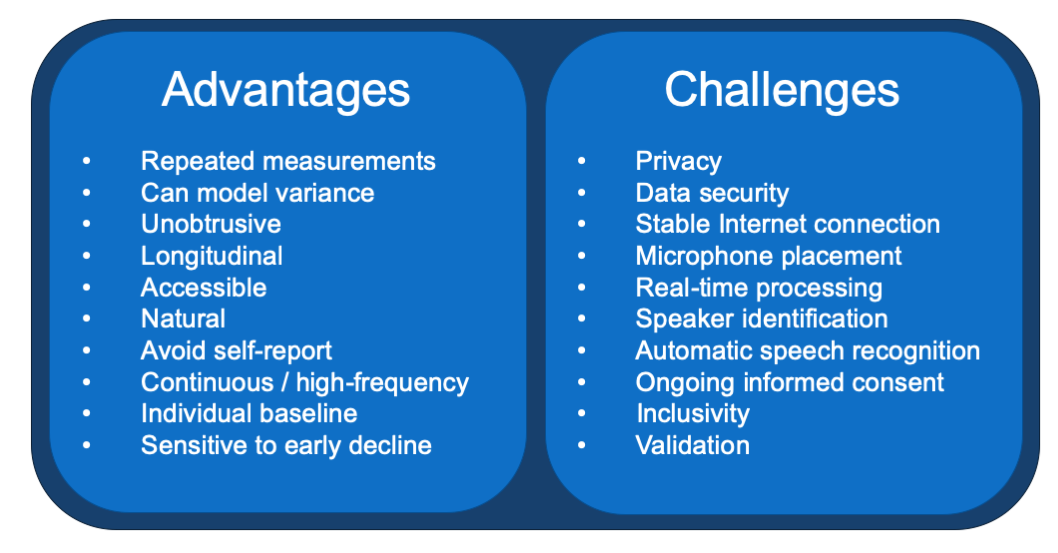}
    \caption{Some of the advantages and challenges associated with smart home speech-based measurement of cognitive status. }
    \label{fig:advantages}
\end{figure*}

Thus far, we have focused on the benefits of smart home speech recording for cognitive monitoring: it can be continuous, unobtrusive, real-time, longitudinal, inclusive to those with mobility issues or who live in remote areas, can avoid unreliable self-reported data, and is potentially more sensitive to decline than traditional cognitive tests \cite{lyons2015pervasive,kaye2008home,kourtis2019digital,nelson2018extending}. However, there are a number of serious challenges which also must be addressed, both on the technical side and from the ethical perspective. These challenges are summarized in Figure~\ref{fig:advantages} and further detailed in the following sections.

\subsection*{Technical Challenges}

There are technical challenges associated with continuously recording, transmitting, and analyzing audio data that may not arise or be addressed in lab-based feasibility studies. 

The recording device itself obviously a critical decision. As Shirmohammadi \textit{et al.} \cite{shirmohammadi2016instrumentation} point out in their review on biomedical instrumentation and measurement, the primary question in home-based biomarker measurement is: ``How can we safeguard that the clinical quality of measurements is achieved using the low cost measurement equipment in a home environment?'' For microphones, the sensitivity, frequency response, and sampling rate will all affect the quality of the audio data. 
%Research has shown that, at least for certain types of acoustic features, smart phone recordings are adequate in comparison to laboratory microphone recordings  \cite{manfredi2017smartphones}.

For recording the data, sensor location relative to the speaker is also important. If using a mobile or wearable device, is the microphone obscured by clothing? If using a virtual assistant or embedded microphones, the distance between speaker and sensor will be constantly changing, and in the case of a robotic assistant, the acoustic properties of the environment will also vary, leading to complications when processing the data.
%\cite{wang2010deconvolutive}. 

Once the data is recorded, it must be transmitted, typically to an encrypted server for processing. Some on-device processing may be possible, but in most cases processing and storage will occur on an external server. One advantage to home-based systems is that they can be connected to wired internet and a continuous power source, avoiding issues due to lost wifi connection or drained batteries. However, any time data is transmitted there is some vulnerability to security threats. Issues of data security, integrity, confidentiality, and authentication must be considered. Secure storage infrastructure is also an issue, as well as the question of how long the audio data must be conserved, and at what resolution. A strategy to protect data against corruption or loss will be crucial for longitudinal analyses spanning many years.

In terms of processing, the details depend on the exact audio features under consideration, but some common challenges remain.

\begin{itemize}
    \item \textbf{Background subtraction:} removing irrelevant ambient noise from the audio stream.
    %, using techniques such as wavelet transform filtering \cite{istrate2006information}. 
    Background speech data, such as from a television or radio, can pose a particular challenge \cite{rabbi2011passive}.
\item \textbf{Sound classification:} speech segments must be distinguished from non-speech segments, for example using pitch detection. %\cite{abu2006security}. 
%\item Noise reduction
\item \textbf{Speaker identification:} the target speaker's voice must be segmented from those of visitors or other household members.
\item \textbf{Automatic speech recognition (ASR):} open-vocabulary ASR of elderly voices in real-world environments has proven to be a challenge. In research studies, the data is often manually transcribed, with the expectation that advances in ASR will be forthcoming \cite{dodge2015social,asgari2017predicting,shinkawa2018word}. Other approaches avoid ASR altogether and focus only on acoustic and prosodic cues \cite{rabbi2011passive}. 
\item \textbf{Feature extraction: } as seen in Figure~\ref{fig:features}, a large number of features may be extracted from the data. The exact feature set may depend on factors such as efficiency of processing, non-redundancy of information, and whether the results need to be computed in real-time. 
\item \textbf{Output:} machine learning and/or time series analysis is used to combine speech features and output a measure of cognitive status, as well as monitor changes over time.
%\cite{yancheva2015using}. 
To date, binary classifiers have often been employed, but more sophisticated approaches are needed. The length of the speech samples may be highly variable and they might have been obtained at irregular time intervals. Appropriate machine learning methods are needed to handle such irregular longitudinal data.
\end{itemize}

Finally, any concerning events that are detected (e.g., a sudden or gradual decline in cognitive ability) must be appropriately communicated to either the user or, preferably, their healthcare provider. Sabbagh \textit{et al.} \cite{sabbagh2020early} refer to this process as ``integration into the clinical care pathway.'' It is of little use, and in fact may be psychologically harmful, if a decline is detected and communicated to the user, but not followed up on by a trained professional.  

\subsection*{Ethical Challenges}

Numerous ethical concerns may be raised by smart home speech recording; the most commonly-raised concern is privacy \cite{ding2011sensor}. People have a right to personal privacy, particularly in their own homes, and systems such as those described above can lead to psychological discomfort, intrusiveness, and the feeling of ``always being watched'' \cite{mittelstadt2017ethics}. There are also questions of data privacy, such as who can access the data \cite{ding2011sensor} and who sees the results of the analysis \cite{sabbagh2020early}. Some of these concerns may be alleviated by technological solutions, such as those described in relation to a  privacy-by-design personal assistant for seniors \cite{seiderer2020development}. Typically, the current approach has been to transmit raw speech samples and process them on a remote server. While this helps with safeguarding the stored data and simplifying the client side of the system, privacy concerns may be allayed if only the extracted features are transmitted to the remote server for generating the desired output. The raw speech data will be discarded after features are extracted locally.

Another concern is that of informed consent. Obviously, it would be inappropriate to install an audio surveillance system in someone's house without their knowledge, but the notion of informed consent becomes trickier when working with users who may have, or develop, dementia. Particularly with the most unobtrusive recording architectures, an individual may forget, over time, to what they have consented \cite{mittelstadt2017ethics}. There is also the issue of visitors to the home: users have indicated that they prefer such a system to be hidden from visitors \cite{ding2011sensor}, but at the same time there is an obligation to disclose to visitors that they may be recorded \cite{mittelstadt2017ethics}. One possible solution to this is using voice detection to not record, or immediately delete, recordings from non-consented speakers \cite{nelson2018extending}.

Concerns have also been raised that using the voice as a digital biomarker may have negative social justice implications. It is well-known that many AI technologies, including speech recognition, perform significantly worse for speakers of languages other than English, people who speak English with an accent, or speakers of African American English (AAE) \cite{geary2020proposed}. Furthermore, installing and maintaining smart home technology can be expensive and therefore not accessible to all. It is important that such technology not perpetuate existing racial and socioeconomic inequities in healthcare access \cite{geary2020proposed}.

\section*{Ways Forward}

There is great potential for speech-based measurement of cognitive status. Many researchers have reported high accuracy in distinguishing between speakers with and without dementia, in cross-sectional studies performed under lab conditions.  Some, but fewer, studies have reported promising results when recording speech \textit{in-situ}, through video chat, wearable devices, and virtual assistants. However, many gaps remain before we can consider this technology fully validated.

Many of the studies described above employed at least some manual processing, either via Wizard-of-Oz dialogues with participants, manual audio segmentation, or manual transcription. A fully automated, end-to-end architecture must be developed, deployed, and evaluated. As previously mentioned, a thorough accounting of the measurement uncertainty at each stage in the pipeline must be performed, as well as an analysis of how those uncertainties propagate through the pipeline \cite{parvis2002medical}.

None of the articles surveyed tracked participants long enough to observe individual trajectories of cognitive change. The ability to do is one of the purported benefits of a home-based system, but it has not yet been achieved in reality, perhaps due to unaddressed privacy concerns. Of course, it makes sense to first design and validate a prototype system over short time spans, but it is our hope to see more long-term, longitudinal investigations of privacy-preserving speech technology in the home. Such a study design would also answer the question of repeatability, or test-retest reliability. Related to this issue, many of the one-shot, cross-sectional research studies on speech analysis actually use elicited speech protocols (e.g., describe a picture, or recall a story). These protocols induce a highly-specific structure and content in the narrative speech. In contrast, most smart home studies measure conversational speech, which is more natural and presumably more repeatable. However, it is not clear whether the same amount or kind of information is available in conversation, as compared to the more targeted elicitation tasks. 

Most studies consider speech as an independent data stream \cite{kourtis2019digital}. Again, this is reasonable first step to assess the advantages and disadvantages of speech as a measurement instrument for cognition, but the true value in a smart home environment will likely lie in multimodal data streams coming together to form a complete assessment of health and well-being. For example, combining speech analysis with other smart home measures of cognition \cite{wallace2017detecting} and physiological health \cite{monteriu2018smart}, in addition to measures of functional ability such as activities of daily living \cite{ando2020advanced} and gait analysis \cite{lahmiri2018gait},  will almost certainly give a more accurate and robust measurement of physical, mental, and cognitive health.

Related to this, most studies based on speech have looked into a single health condition;
%for example, assessing cognitive decline \cite{gosztolya2019identifying}, mental well-being \cite{rabbi2011passive} or some respiratory conditions \cite{porter2019prospective} among others. 
a holistic approach that simultaneously assesses multiple health conditions may help justify privacy concerns and facilitates integration of such a system within a smart home environment.

Finally, there is a pressing need for validation with more diverse user populations. In additional to clinical diversity (\textit{i.e.}, different types of dementia and related pathologies, brain injuries, comorbid mental health issues, \textit{etc.}), it is important that these technologies are validated with users representing different cultural and educational backgrounds, varieties of English, and varied living situations. This will ensure the validity and  reproducibility of the instrument on a heterogeneous user base.

\section*{Conclusion}

Smart home technologies offer the potential to change the way we live throughout our lives. Speech, as a natural and ubiquitous human behavior, will undoubtedly be incorporated into such environments -- and indeed, already is, through our use of voice commands and virtual assistants. As a measurement instrument of cognitive status, speech presents many opportunities, but also many open questions to be answered in future research.

It will be crucial to bring together multi-disciplinary teams to fully understand and serve the needs of older adults as they face the challenges of cognitive decline. Nelson \textit{et al.} \cite{nelson2018extending} recommend recruiting expertise from disciplines as varied as  ``behavioral science, affective computing, mobile sensing, mental health, signal processing, artificial intelligence, biomedical engineering, data mining, computer networks, machine learning, bioethics, and technology development, along with industry partners.'' In this article we have emphasized the need for the thoughtful application of the principles of measurement to the assessment of such automated systems, to protect and promote the health and well-being of our aging population. 

\bibliographystyle{IEEEtran}
\bibliography{references}

\begin{thebibliography}{10}
\providecommand{\url}[1]{#1}
\csname url@samestyle\endcsname
\providecommand{\newblock}{\relax}
\providecommand{\bibinfo}[2]{#2}
\providecommand{\BIBentrySTDinterwordspacing}{\spaceskip=0pt\relax}
\providecommand{\BIBentryALTinterwordstretchfactor}{4}
\providecommand{\BIBentryALTinterwordspacing}{\spaceskip=\fontdimen2\font plus
\BIBentryALTinterwordstretchfactor\fontdimen3\font minus
  \fontdimen4\font\relax}
\providecommand{\BIBforeignlanguage}[2]{{%
\expandafter\ifx\csname l@#1\endcsname\relax
\typeout{** WARNING: IEEEtran.bst: No hyphenation pattern has been}%
\typeout{** loaded for the language `#1'. Using the pattern for}%
\typeout{** the default language instead.}%
\else
\language=\csname l@#1\endcsname
\fi
#2}}
\providecommand{\BIBdecl}{\relax}
\BIBdecl

\bibitem{anderson2017technology}
M.~Anderson and A.~Perrin, ``Technology use among seniors,'' \emph{Pew Research
  Center for Internet \& Technology}, 2017.

\bibitem{lyons2015pervasive}
B.~E. Lyons, D.~Austin, A.~Seelye, J.~Petersen, J.~Yeargers, T.~Riley,
  N.~Sharma, N.~Mattek, K.~Wild, H.~Dodge \emph{et~al.}, ``Pervasive computing
  technologies to continuously assess {A}lzheimer’s disease progression and
  intervention efficacy,'' \emph{Frontiers in Aging Neuroscience}, vol.~7, p.
  102, 2015.

\bibitem{fraser2016linguistic}
K.~C. Fraser, J.~A. Meltzer, and F.~Rudzicz, ``Linguistic features identify
  {A}lzheimer's disease in narrative speech,'' \emph{Journal of Alzheimer's
  Disease}, vol.~49, no.~2, pp. 407--422, 2016.

\bibitem{mirheidari2016diagnosing}
B.~Mirheidari, D.~Blackburn, M.~Reuber, T.~Walker, and H.~Christensen,
  ``Diagnosing people with dementia using automatic conversation analysis,'' in
  \emph{Proceedings of Interspeech}.\hskip 1em plus 0.5em minus 0.4em\relax
  ISCA, 2016, pp. 1220--1224.

\bibitem{roark2011spoken}
B.~Roark, M.~Mitchell, J.-P. Hosom, K.~Hollingshead, and J.~Kaye, ``Spoken
  language derived measures for detecting mild cognitive impairment,''
  \emph{IEEE Transactions on Audio, Speech, and Language Processing}, vol.~19,
  no.~7, pp. 2081--2090, 2011.

\bibitem{beltrami2018speech}
D.~Beltrami, G.~Gagliardi, R.~Rossini~Favretti, E.~Ghidoni, F.~Tamburini, and
  L.~Calz{\`a}, ``Speech analysis by natural language processing techniques:
  {A} possible tool for very early detection of cognitive decline?''
  \emph{Frontiers in Aging Neuroscience}, vol.~10, p. 369, 2018.

\bibitem{gosztolya2019identifying}
G.~Gosztolya, V.~Vincze, L.~T{\'o}th, M.~P{\'a}k{\'a}ski, J.~K{\'a}lm{\'a}n,
  and I.~Hoffmann, ``Identifying mild cognitive impairment and mild
  {A}lzheimer’s disease based on spontaneous speech using {ASR} and
  linguistic features,'' \emph{Computer Speech \& Language}, vol.~53, pp.
  181--197, 2019.

\bibitem{komeili2019talk2me}
M.~Komeili, C.~Pou-Prom, D.~Liaqat, K.~C. Fraser, M.~Yancheva, and F.~Rudzicz,
  ``Talk2me: Automated linguistic data collection for personal assessment,''
  \emph{Plos one}, vol.~14, no.~3, p. e0212342, 2019.

\bibitem{parvis2002medical}
M.~Parvis and A.~Vallan, ``Medical measurements and uncertainties,'' \emph{IEEE
  Instrumentation \& Measurement Magazine}, vol.~5, no.~2, pp. 12--17, 2002.

\bibitem{crocco2016audio}
\BIBentryALTinterwordspacing
M.~Crocco, M.~Cristani, A.~Trucco, and V.~Murino, ``Audio surveillance: {A}
  systematic review,'' \emph{ACM Computing Surveys}, vol.~48, no.~4, Feb. 2016.
  [Online]. Available: \url{https://doi.org/10.1145/2871183}
\BIBentrySTDinterwordspacing

\bibitem{ferrero2006measurement}
A.~Ferrero and S.~Salicone, ``Measurement uncertainty,'' \emph{IEEE
  Instrumentation \& Measurement Magazine}, vol.~9, no.~3, pp. 44--51, 2006.

\bibitem{berisha2017longitudinal}
V.~Berisha, S.~Wang, A.~LaCross, J.~Liss, and P.~Garcia-Filion, ``Longitudinal
  changes in linguistic complexity among professional football players,''
  \emph{Brain and language}, vol. 169, pp. 57--63, 2017.

\bibitem{kaye2008home}
J.~Kaye, ``Home-based technologies: A new paradigm for conducting dementia
  prevention trials,'' \emph{Alzheimer's \& Dementia}, vol.~4, no.~1, pp. S60
  -- S66, 2008.

\bibitem{kaye2011intelligent}
J.~A. Kaye, S.~A. Maxwell, N.~Mattek, T.~L. Hayes, H.~Dodge, M.~Pavel, H.~B.
  Jimison, K.~Wild, L.~Boise, and T.~A. Zitzelberger, ``Intelligent systems for
  assessing aging changes: {H}ome-based, unobtrusive, and continuous assessment
  of aging,'' \emph{Journals of Gerontology Series B: Psychological Sciences
  and Social Sciences}, vol.~66, no. suppl\_1, pp. i180--i190, 2011.

\bibitem{dodge2015social}
H.~H~Dodge, N.~Mattek, M.~Gregor, M.~Bowman, A.~Seelye, O.~Ybarra, M.~Asgari,
  and J.~A~Kaye, ``Social markers of mild cognitive impairment: {P}roportion of
  word counts in free conversational speech,'' \emph{Current Alzheimer
  Research}, vol.~12, no.~6, pp. 513--519, 2015.

\bibitem{asgari2017predicting}
M.~Asgari, J.~Kaye, and H.~Dodge, ``Predicting mild cognitive impairment from
  spontaneous spoken utterances,'' \emph{Alzheimer's \& Dementia: Translational
  Research \& Clinical Interventions}, vol.~3, no.~2, pp. 219--228, 2017.

\bibitem{bowers20183cap}
K.~M. Bowers, R.~H. Hariri, and K.~A. Price, ``{3CAP:} {C}ategorizing the
  cognitive capabilities of {A}lzheimer's patients in a smart home
  environment,'' in \emph{Proceedings of the 4th ACM SIGSOFT International
  Workshop on NLP for Software Engineering}, 2018, pp. 34--37.

\bibitem{chen2019developing}
\BIBentryALTinterwordspacing
R.~Chen, F.~Jankovic, N.~Marinsek, L.~Foschini, L.~Kourtis, A.~Signorini,
  M.~Pugh, J.~Shen, R.~Yaari, V.~Maljkovic, M.~Sunga, H.~H. Song, H.~J. Jung,
  B.~Tseng, and A.~Trister, ``Developing measures of cognitive impairment in
  the real world from consumer-grade multimodal sensor streams,'' in
  \emph{Proceedings of the 25TH ACM SIGKDD Conference on Knowledge Discovery
  and Data Mining (KDD)}, 2019. [Online]. Available:
  \url{https://dl.acm.org/doi/pdf/10.1145/3292500.3330690}
\BIBentrySTDinterwordspacing

\bibitem{sabbagh2020early}
M.~N. Sabbagh, M.~Boada, S.~Borson, P.~M. Doraiswamy, B.~Dubois, J.~Ingram,
  A.~Iwata, A.~Porsteinsson, K.~Possin, G.~Rabinovici \emph{et~al.}, ``Early
  detection of mild cognitive impairment ({MCI}) in an at-home setting,''
  \emph{The Journal of Prevention of Alzheimer's Disease}, pp. 1--8, 2020.

\bibitem{netscher2016applications}
G.~Netscher, ``Applications of machine learning to support dementia care
  through commercially available off-the-shelf sensing,'' \emph{Berkeley:
  University of California}, 2016.

\bibitem{shinkawa2018word}
K.~Shinkawa and Y.~Yamada, ``Word repetition in separate conversations for
  detecting dementia: {A} preliminary evaluation on data of regular monitoring
  service,'' \emph{AMIA Summits on Translational Science Proceedings}, vol.
  2018, p. 206, 2018.

\bibitem{kourtis2019digital}
L.~C. Kourtis, O.~B. Regele, J.~M. Wright, and G.~B. Jones, ``Digital
  biomarkers for {A}lzheimer's disease: {T}he mobile/wearable devices
  opportunity,'' \emph{NPJ Digital Medicine}, vol.~2, no.~1, pp. 1--9, 2019.

\bibitem{rabbi2011passive}
M.~Rabbi, S.~Ali, T.~Choudhury, and E.~Berke, ``Passive and in-situ assessment
  of mental and physical well-being using mobile sensors,'' in
  \emph{Proceedings of the 13th International Conference on Ubiquitous
  Computing}, 2011, pp. 385--394.

\bibitem{polsinelli2020natural}
A.~J. Polsinelli, S.~A. Moseley, M.~D. Grilli, E.~L. Glisky, and M.~R. Mehl,
  ``Natural, everyday language use provides a window into the integrity of
  older adults' executive functioning,'' \emph{The Journals of Gerontology:
  Series B}, 2020.

\bibitem{kobayashi2019effects}
M.~Kobayashi, A.~Kosugi, H.~Takagi, M.~Nemoto, K.~Nemoto, T.~Arai, and
  Y.~Yamada, ``Effects of age-related cognitive decline on elderly user
  interactions with voice-based dialogue systems,'' in \emph{IFIP Conference on
  Human-Computer Interaction}.\hskip 1em plus 0.5em minus 0.4em\relax Springer,
  2019, pp. 53--74.

\bibitem{yamada2020predicting}
Y.~Yamada, K.~Shinkawa, A.~Kosugi, M.~Kobayashi, H.~Takagi, M.~Nemoto,
  K.~Nemoto, and T.~Arai, ``Predicting future accident risks of older drivers
  by speech data from a voice-based dialogue system: {A} preliminary result,''
  in \emph{International Conference on Applied Human Factors and
  Ergonomics}.\hskip 1em plus 0.5em minus 0.4em\relax Springer, 2020, pp.
  131--137.

\bibitem{walker2020characterising}
G.~Walker, L.-A. Morris, H.~Christensen, B.~Mirheidari, M.~Reuber, and D.~J.
  Blackburn, ``Characterising spoken responses to an intelligent virtual agent
  by persons with mild cognitive impairment,'' \emph{Clinical Linguistics \&
  Phonetics}, pp. 1--16, 2020.

\bibitem{pou2020conversational}
C.~Pou-Prom, S.~Raimondo, and F.~Rudzicz, ``A conversational robot for older
  adults with {A}lzheimer's disease,'' \emph{ACM Transactions on Human-Robot
  Interaction (THRI)}, vol.~9, no.~3, pp. 1--25, 2020.

\bibitem{portet2013design}
F.~Portet, M.~Vacher, C.~Golanski, C.~Roux, and B.~Meillon, ``Design and
  evaluation of a smart home voice interface for the elderly: {A}cceptability
  and objection aspects,'' \emph{Personal and Ubiquitous Computing}, vol.~17,
  no.~1, pp. 127--144, 2013.

\bibitem{amiribesheli2018tailored}
M.~Amiribesheli and H.~Bouchachia, ``A tailored smart home for dementia care,''
  \emph{Journal of Ambient Intelligence and Humanized Computing}, vol.~9,
  no.~6, pp. 1755--1782, 2018.

\bibitem{nelson2018extending}
B.~W. Nelson and N.~B. Allen, ``Extending the passive-sensing toolbox: {U}sing
  smart-home technology in psychological science,'' \emph{Perspectives on
  Psychological Science}, vol.~13, no.~6, pp. 718--733, 2018.

\bibitem{shirmohammadi2016instrumentation}
S.~Shirmohammadi, K.~Barbe, D.~Grimaldi, S.~Rapuano, and S.~Grassini,
  ``Instrumentation and measurement in medical, biomedical, and healthcare
  systems,'' \emph{IEEE Instrumentation \& Measurement Magazine}, vol.~19,
  no.~5, pp. 6--12, 2016.

\bibitem{ding2011sensor}
D.~Ding, R.~A. Cooper, P.~F. Pasquina, and L.~Fici-Pasquina, ``Sensor
  technology for smart homes,'' \emph{Maturitas}, vol.~69, no.~2, pp. 131--136,
  2011.

\bibitem{mittelstadt2017ethics}
B.~Mittelstadt, ``Ethics of the health-related internet of things: {A}
  narrative review,'' \emph{Ethics and Information Technology}, vol.~19, no.~3,
  pp. 157--175, 2017.

\bibitem{seiderer2020development}
A.~Seiderer, H.~Ritschel, and E.~Andr{\'e}, ``Development of a
  privacy-by-design speech assistant providing nutrient information for
  {G}erman seniors,'' in \emph{Proceedings of the 6th EAI International
  Conference on Smart Objects and Technologies for Social Good}, 2020, pp.
  114--119.

\bibitem{geary2020proposed}
A.~Geary-Teeter and R.~Hosseini~Ghomi, ``A proposed framework for addressing
  social justice concerns in future digital biomarker research,'' 2020.

\bibitem{wallace2017detecting}
B.~Wallace, F.~Knoefel, R.~Goubran, P.~Masson, A.~Baker, B.~Allard, V.~Guana,
  and E.~Stroulia, ``Detecting cognitive ability changes in patients with
  moderate dementia using a modified ``{Whack-a-Mole}'' game,'' \emph{IEEE
  Transactions on Instrumentation and Measurement}, vol.~67, no.~7, pp.
  1521--1534, 2017.

\bibitem{monteriu2018smart}
A.~Monteri{\`u}, M.~R. Prist, E.~Frontoni, S.~Longhi, F.~Pietroni, S.~Casaccia,
  L.~Scalise, A.~Cenci, L.~Romeo, R.~Berta \emph{et~al.}, ``A smart sensing
  architecture for domestic monitoring: {M}ethodological approach and
  experimental validation,'' \emph{Sensors}, vol.~18, no.~7, p. 2310, 2018.

\bibitem{ando2020advanced}
B.~And{\`o}, S.~Baglio, S.~Castorina, R.~Crispino, and V.~Marietta, ``Advanced
  sensing solutions for ambient assisted living: {T}he {NATIFLife} framework,''
  \emph{IEEE Instrumentation \& Measurement Magazine}, vol.~23, no.~4, pp.
  33--40, 2020.

\bibitem{lahmiri2018gait}
S.~Lahmiri, ``Gait nonlinear patterns related to {P}arkinson's disease and
  age,'' \emph{IEEE Transactions on Instrumentation and Measurement}, vol.~68,
  no.~7, pp. 2545--2551, 2018.

\end{thebibliography}
%\begin{thebibliography}{1}


%\end{thebibliography}


%\begin{IEEEbiography}{Kathleen C. Fraser}
%\end{IEEEbiography}

%\begin{IEEEbiography}{Author}
%\end{IEEEbiography}


\newpage

\noindent \textbf{Biographies}

\bigskip

\noindent \textbf{Kathleen C. Fraser} received her Ph.D.\@ in Computer Science from the University of Toronto. She held a post-doctoral position at the University of Gothenburg and currently works as a researcher at the National Research Council Canada. Her research expertise lies in natural language processing and machine learning, particularly as they apply to health and well-being in aging. 

\bigskip

\noindent \textbf{Majid Komeili} received his Ph.D. degree in Electrical and Computer Engineering from the University of Toronto, Toronto, ON, Canada in 2017. He is currently an Assistant Professor in the School of Computer Science at Carleton University, Ottawa, ON, Canada. He performs fundamental and applied research in machine learning.


%\vfill


\end{document}